\documentclass[sigconf]{acmart}

\AtBeginDocument{%
  }

\copyrightyear{2023}
\acmYear{2023}
\setcopyright{rightsretained}
\acmConference[MM '23]{Proceedings of the 31st ACM International Conference on Multimedia}{October 29-November 3, 2023}{Ottawa, ON, Canada}
\acmBooktitle{Proceedings of the 31st ACM International Conference on Multimedia (MM '23), October 29-November 3, 2023, Ottawa, ON, Canada}
\acmDOI{10.1145/3581783.3612836}
\acmISBN{979-8-4007-0108-5/23/10}

\usepackage{wrapfig}
\usepackage{multirow}
\usepackage{subfigure}
\usepackage[ruled, linesnumbered]{algorithm2e}

\makeatletter
\gdef\@copyrightpermission{
	\begin{minipage}{0.3\columnwidth}
		\href{https://creativecommons.org/licenses/by/4.0/}{\includegraphics[width=0.90\textwidth]{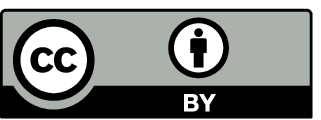}}
	\end{minipage}\hfill
	\begin{minipage}{0.7\columnwidth}
		\href{https://creativecommons.org/licenses/by/4.0/}{This work is licensed under a Creative Commons Attribution International 4.0 License.}
	\end{minipage}
	\vspace{5pt}
}
\makeatother

\begin{document}
	
	\title{MER 2023: Multi-label Learning, Modality Robustness, and Semi-Supervised Learning}
	
	\author{Zheng Lian}
	\affiliation{%
		\institution{Institute of Automation, Chinese Academy of Sciences (CAS)}
		\city{Beijing}
		\country{China}
	}

	\author{Haiyang Sun}
	\affiliation{%
		\institution{University of Chinese Academy \\ of Sciences}
		\city{Beijing}
		\country{China}
	}
	
	\author{Licai Sun}
	\affiliation{%
		\institution{University of Chinese Academy \\ of Sciences}
		\city{Beijing}
		\country{China}
	}
	
	\author{Kang Chen}
	\affiliation{%
		\institution{Peking University}
		\city{Beijing}
		\country{China}
	}

	\author{Mingyu Xu}
	\affiliation{%
		\institution{Institute of Automation, CAS}
		\city{Beijing}
		\country{China}
	}
	
	\author{Kexin Wang}
	\affiliation{%
		\institution{Institute of Automation, CAS}
		\city{Beijing}
		\country{China}
	}

	\author{Ke Xu}
	\affiliation{%
		\institution{University of Chinese Academy \\ of Sciences}
		\city{Beijing}
		\country{China}
	}
	
	\author{Yu He}
	\affiliation{%
		\institution{University of Chinese Academy \\ of Sciences}
		\city{Beijing}
		\country{China}
	}
	
	\author{Ying Li}
	\affiliation{%
		\institution{Shandong Normal University}
		\city{Shandong}
		\country{China}
	}
	
	\author{Jinming Zhao}
	\affiliation{%
		\institution{Renmin University of China}
		\city{Beijing}
		\country{China}
	}
	
	\author{Ye Liu}
	\affiliation{%
		\institution{Institute of Psychology, CAS}
		\city{Beijing}
		\country{China}
	}
	
	\author{Bin Liu}
	\affiliation{%
		\institution{Institute of Automation, CAS}
		\city{Beijing}
		\country{China}
	}
	
	\author{Jiangyan Yi}
	\affiliation{%
		\institution{Institute of Automation, CAS}
		\city{Beijing}
		\country{China}
	}
	
	\author{Meng Wang}
	\affiliation{%
		\institution{Ant Group}
		\city{Beijing}
		\country{China}
	}

	\author{Erik Cambria}
	\affiliation{%
		\institution{Nanyang Technological University}
		\country{Singapore}
	}
	
	\author{Guoying Zhao}
	\affiliation{%
		\institution{University of Oulu}
		\city{Oulu}
		\country{Finland}
	}
	
	\author{Björn W. Schuller}
	\affiliation{%
		\institution{Imperial College London}
		\city{London}
		\country{United Kingdom}
	}

	\author{Jianhua Tao}
	\affiliation{%
		\institution{Tsinghua University}
		\city{Beijing}
		\country{China}
	}

	\renewcommand{\shortauthors}{Zheng Lian et al.}
	
	\begin{abstract}
	The first Multimodal Emotion Recognition Challenge (MER 2023)\footnote{\emph{http://merchallenge.cn/mer2023}} was successfully held at ACM Multimedia. The challenge focuses on system robustness and consists of three distinct tracks: (1) MER-MULTI, where participants are required to recognize both discrete and dimensional emotions; (2) MER-NOISE, in which noise is added to test videos for modality robustness evaluation; (3) MER-SEMI, which provides a large amount of unlabeled samples for semi-supervised learning. In this paper, we introduce the motivation behind this challenge, describe the benchmark dataset, and provide some statistics about participants. To continue using this dataset after MER 2023, please sign a new End User License Agreement\footnote{\emph{https://drive.google.com/file/d/1LOW2e6ZuyUjurVF0SNPisqSh4VzEl5lN}} and send it to our official email address\footnote{\emph{merchallenge.contact@gmail.com}}. We believe this high-quality dataset can become a new benchmark in multimodal emotion recognition, especially for the Chinese research community.
	\end{abstract}

	
	
	\maketitle

	\section{Introduction}
	Multimodal emotion recognition has become an important research topic due to its wide-ranging applications in human-computer interaction. Over the past few decades, researchers have proposed various approaches \cite{ebrahimi2015recurrent, kahou2016emonets, young2018recent}. But due to their low robustness in complex environments, existing techniques do not fully meet the demands in practice. To this end, we launch a Multimodal Emotion Recognition Challenge (MER 2023), which aims to improve system robustness from three aspects: multi-label learning, modality robustness, and semi-supervised learning.
	
	Annotating with both discrete and dimensional emotions is common in current datasets \cite{busso2008iemocap, zadeh2018multimodal}. Existing works mainly utilize multi-task learning to predict all labels simultaneously \cite{chen2017multimodal2, akhtar2019multi}. However, these works ignore the correlation between discrete and dimensional emotions. For example, valence is a dimensional emotion that reflects the degree of pleasure. For negative emotions (such as \emph{anger} and \emph{sadness}), the valence score should be less than 0; for positive emotions (such as \emph{happiness}), the valence score should be greater than 0. To fully exploit the multi-label correlation, we launch the MER-MULTI sub-challenge, which encourages participants to exploit the appropriate loss function \cite{chou2022exploiting} or model structure \cite{wang2022emotional} to boost recognition performance. 
	
	Many factors may lead to modality perturbation, which increases the difficulty of emotion recognition. Recently, researchers have proposed various strategies to deal with this problem \cite{zhao2021missing, yuan2021transformer, sun2022efficient}. But due to the lack of benchmark datasets, existing works mainly rely on their own simulated missing conditions to evaluate modality robustness. To this end, we launch the MER-NOISE sub-challenge, which provides a benchmark test set focusing on more realistic modality perturbations such as background noise and blurry videos. In this sub-challenge, we encourage participants to use data augmentation \cite{hazarika2022analyzing} or other more advanced techniques \cite{zhang2022deep, lian2023gcnet}.
	
	Meanwhile, it is difficult to collect large amounts of emotion-labeled samples due to the high annotation cost. Training with limited data harms the generalization ability of recognition systems. To address this issue, researchers have exploited various pre-trained models for video emotion recognition \cite{lian2019unsupervised, mao2022biases}. However, task similarity impacts the performance of transfer learning \cite{weiss2016survey}. Existing video-level pre-trained models mainly focus on action recognition rather than expression videos \cite{tong2022videomae}. In this paper, we extract human-centered video clips from movies and TV series that contain emotional expressions. We then launch the MER-SEMI sub-challenge, encouraging participants to use semi-supervised learning \cite{he2022masked, tong2022videomae} to achieve better performance.
	
	Therefore, MER 2023 consists of three sub-challenges: MER-MULTI, MER-NOISE, and MER-SEMI. Different from existing challenges (such as AVEC \cite{schuller2011avec, schuller2012avec, valstar2013avec, valstar2014avec, ringeval2015avec, valstar2016avec, ringeval2017avec, ringeval2018avec, ringeval2019avec}, EmotiW \cite{dhall2013emotion, dhall2014emotion, dhall2015video, dhall2016emotiw, dhall2017individual, dhall2018emotiw, dhall2019emotiw, dhall2020emotiw}, and MuSE \cite{stappen2020muse, stappen2021muse, amiriparian2022muse}), we mainly focus on system robustness, and provide a common platform and benchmark test sets for performance evaluation. We plan to organize a series of challenges and related workshops that bring together researchers from all over the world to discuss recent research and future directions in this field.
	
	\begin{table}[t]
		\centering
		\caption{Statistical information for the challenge dataset (duration: hh:mm:ss).}
		\label{Table1}
		\begin{tabular}{l|cc|c}
			\hline
			\multirow{2}{*}{Partition} & \multicolumn{2}{c|}{\# of samples} & \multirow{2}{*}{Duration} \\
			& labeled & unlabeled & \\
			\hline \hline
			Train$\&$Val 	& 3373 	& 0 	& 03:45:47 \\ 
			\hline
			MER-MULTI 		& 411 	& 0 	& 00:28:09 \\
			MER-NOISE 		& 412 	& 0 	& 00:26:23 \\
			MER-SEMI 		& 834 	& 73148 & 67:41:24 \\
			\hline
		\end{tabular}
	\end{table}
	
	\begin{figure}[t]
		\centering
		\includegraphics[width=0.66\linewidth, trim=120 60 10 20]{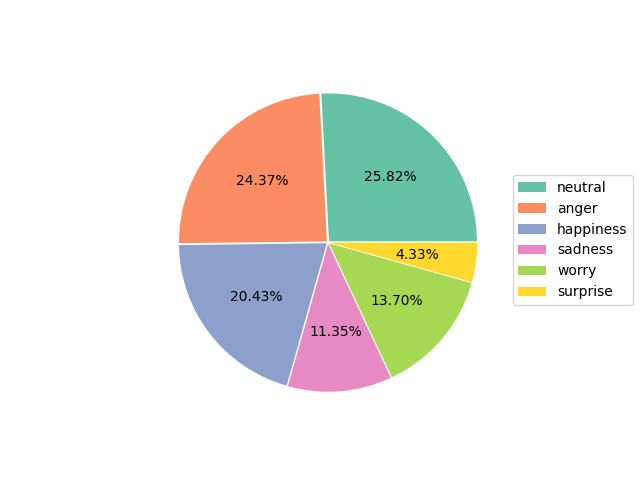}
		\caption{Distribution of discrete emotions (Train$\&$Val).}
		\label{Figure2}
	\end{figure}

	\begin{figure}[t]
		\begin{center}
			\subfigure[neutral]{
				\label{Figure3-1}
				\centering
				\includegraphics[width=0.30\linewidth, trim=15 0 15 0]{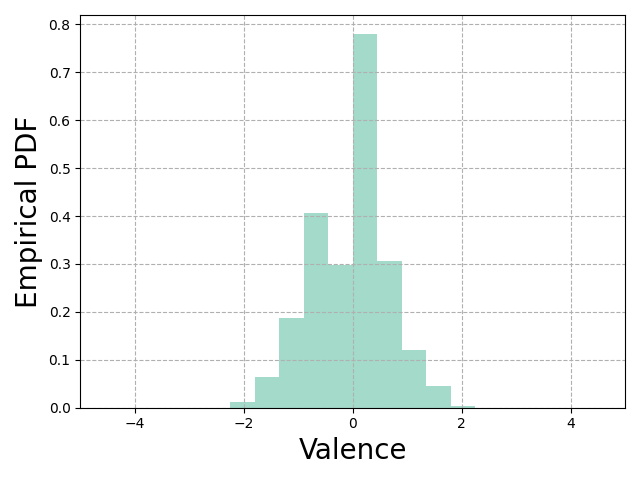}
			} 
			\subfigure[anger]{
				\label{Figure3-2}
				\centering
				\includegraphics[width=0.30\linewidth, trim=15 0 15 0]{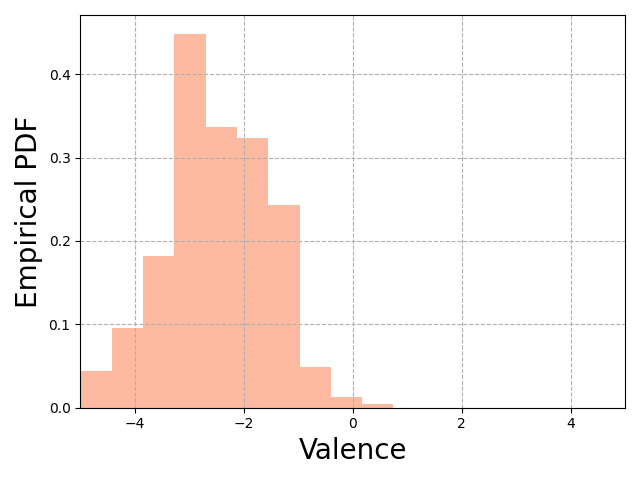}
			}
			\subfigure[happiness]{
				\label{Figure3-3}
				\centering
				\includegraphics[width=0.30\linewidth, trim=15 0 15 0]{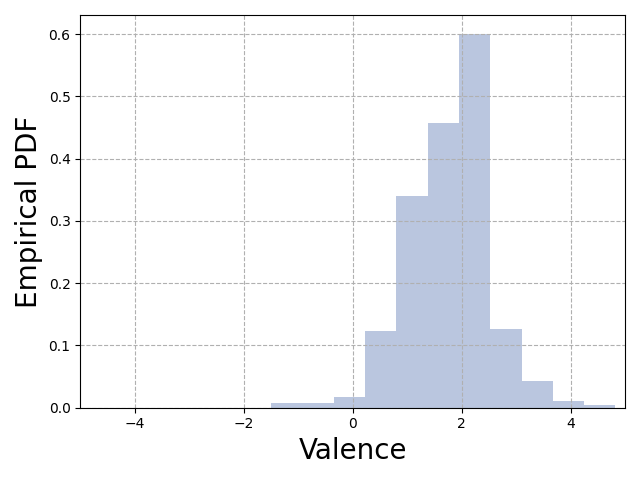}
			} 
			
			\subfigure[sadness]{
				\label{Figure3-4}
				\centering
				\includegraphics[width=0.3\linewidth, trim=15 0 15 0]{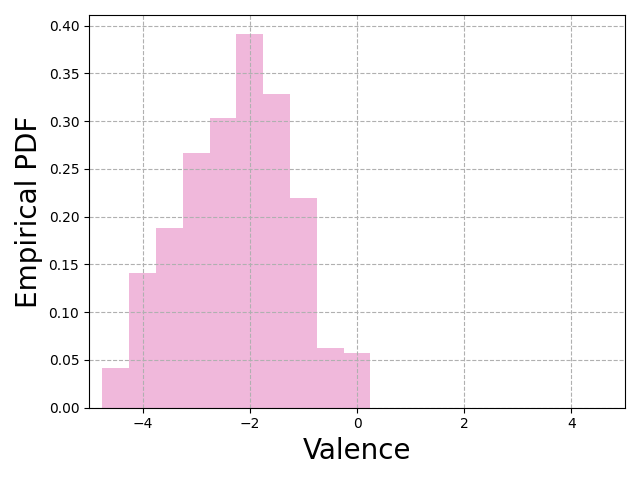}
			}
			\subfigure[worry]{
				\label{Figure3-5}
				\centering
				\includegraphics[width=0.3\linewidth, trim=15 0 15 0]{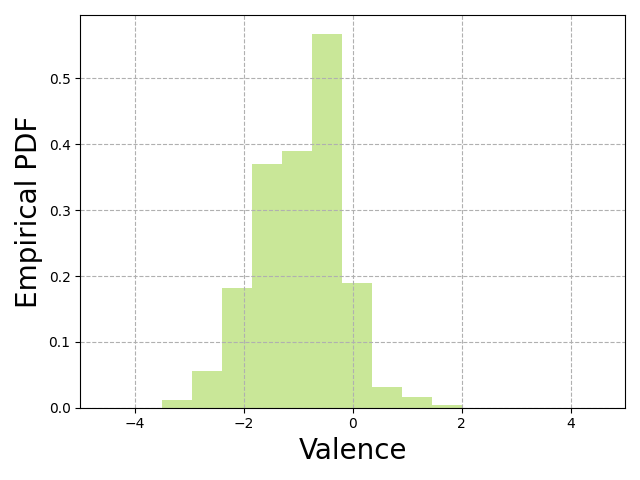}
			}
			\subfigure[surprise]{
				\label{Figure3-6}
				\centering
				\includegraphics[width=0.3\linewidth, trim=15 0 15 0]{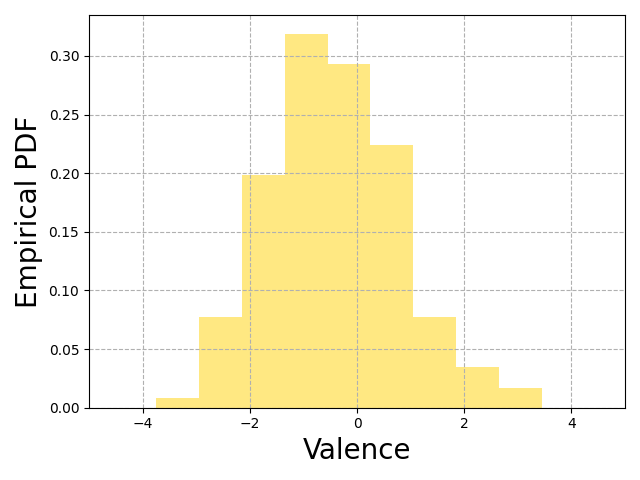}
			}
			
		\end{center}
		\caption{Empirical PDF on the valence for different discrete emotions (Train$\&$Val).}
		\label{Figure3}
	\end{figure}

	\section{Challenge Dataset}
	MER 2023 employs an extended version of CHEAVD for performance evaluation. Due to the small size of CHEAVD, we implement a fully automatic strategy to collect large amounts of unlabeled video clips; due to the low annotation consistency of CHEAVD, we adopt a stricter data selection approach and split the dataset into reliable and unreliable parts. As for reliable samples, we further divide them into three subsets: Train$\&$Val, MER-MULTI, and MER-NOISE. As for unreliable samples, we treat them as unlabeled data and merge them with automatically-collected samples to form MER-SEMI. Statistics of each subset are shown in Table \ref{Table1}.
	
	Figure \ref{Figure2} summarizes the distribution of discrete emotions. Despite some imbalance, our dataset still exhibits a relatively high balance compared to other mainstream benchmarks such as MELD \cite{poria2019meld} and CMU-MOSEI \cite{zadeh2018multimodal}. Figure \ref{Figure3} further reveals the relationship between discrete emotions and valences. Valence serves as an indicator of pleasure, and the value from small to large means the sentiment from negative to positive. From this figure, we observe that the valence distribution of different discrete labels is quite reasonable. Negative emotions (such as \emph{anger}, \emph{sadness}, and \emph{worry}) predominantly exhibit valences below 0. Conversely, positive emotions (such as \emph{happiness}) primarily exhibit valences above 0. The valence associated with \emph{neutral} centers around 0. Notably, \emph{surprise} is a fairly complex emotion that contains multiple meanings such as sadly surprised, angrily surprised, or happily surprised. Hence, its valence ranges from negative to positive. These findings ensure the high quality of our labels and demonstrate the necessity of incorporating both discrete and dimensional annotations, as they can help us distinguish some subtle differences in emotional states.
	
	To download the dataset, participants should fill out an End User License Agreement (EULA)\footnote{\emph{https://drive.google.com/file/d/1I0ZuPzL96W4Ow3KdF3N6D-mDhS5qJc-x}}, which requires participants to use this dataset only for academic research and not to edit or upload samples to the Internet. For each track, participants can submit 20 times per day with a maximum of 200 times: MER-MULTI\footnote{\emph{https://codalab.lisn.upsaclay.fr/competitions/14164}}, MER-NOISE\footnote{\emph{https://codalab.lisn.upsaclay.fr/competitions/14165}}, and MER-SEMI\footnote{\emph{https://codalab.lisn.upsaclay.fr/competitions/14166}}. At the end of the challenge, each team is required to submit a paper describing their approach. For each paper, the program committee will conduct a double-blind review of the scientific quality, novelty, and technical quality. To continue using this dataset after the challenge, please sign a new EULA\footnote{\emph{https://drive.google.com/file/d/1LOW2e6ZuyUjurVF0SNPisqSh4VzEl5lN}} and send it to our official email address\footnote{\emph{merchallenge.contact@gmail.com}}. We will provide the test set labels to facilitate further usage. We believe this dataset can serve as a new benchmark in robust multimodal emotion recognition, especially for the Chinese research community.
	
	\begin{table}[t]
		\centering
		\renewcommand\tabcolsep{6pt}
		\caption{MER-MULTI leaderboard.}
		\label{Table2}
		\begin{tabular}{l|l|c}
			\hline
			Rank & Team & Combined $(\uparrow)$  \\
			\hline
			1 & sense-dl-lab & 0.7005 \\
			2 & AIPL-BME-SEU & 0.6860 \\
			3 & USTC-qw & 0.6846 \\
			4 & AI4AI & 0.6783 \\
			5 & T\_MERG & 0.6765 \\
			6 & SZTU-MIPS & 0.6702 \\
			7 & Desheng & 0.6675 \\
			8 & Suda\_iai & 0.6087 \\
			9 & Emotion recognition group & 0.6025 \\
			10 & SUST-EiAi-Team & 0.5988 \\
			11 & FudanDML & 0.5880 \\
			12 & MCI-SCUT & 0.5819 \\
			13 & Beihang University & 0.5713 \\
			14 & ADDD & 0.5673 \\
			15 & Winner & 0.5655 \\
			-- & \textbf{Baseline} & \textbf{0.56} \\
			16 & TUA1 & 0.5561 \\
			17 & SCUTer & 0.5551 \\
			18 & CiL Fighting! & 0.5453 \\
			19 & CCNUNLP & 0.5446 \\
			20 & Quaint Critters & 0.5422 \\
			21 & Emo.avi & 0.4977 \\
			22 & SDNU\_AIASC & 0.4639 \\
			23 & Cognitist & 0.1612 \\
			\hline
		\end{tabular}
	\end{table}

	\begin{table}[t]
		\centering
		\renewcommand\tabcolsep{8pt}
		\caption{MER-NOISE leaderboard.}
		\label{Table3}
		\begin{tabular}{l|l|c}
			\hline
			Rank & Team & Combined $(\uparrow)$ \\
			\hline
			1 & sense-dl-lab & {0.6846} \\
			2 & AIPL-BME-SEU & 0.6694 \\
			3 & AI4AI & 0.6371 \\
			4 & LeVoice & 0.6256 \\
			5 & SZTU-MIPS & 0.6247 \\
			6 & USTC-qw & 0.6162 \\
			7 & Voice of Soul & 0.6140 \\
			8 & Desheng & 0.5707 \\
			9 & Beihang University & 0.5462 \\
			10 & SUST-EiAi-Team & 0.5455 \\
			11 & Triple Six & 0.5444 \\
			12 & FudanDML & 0.5378 \\
			13 & Delta & 0.5339 \\
			14 & USTBJDL822 & 0.5075 \\
			15 & MCI-SCUT & 0.5003 \\
			16 & Trailblazers & 0.4669 \\
			-- & \textbf{Baseline} & \textbf{0.41} \\
			17 & ADDD & 0.4055 \\
			18 & CiL Fighting! & 0.3863 \\
			19 & Suda\_iai & 0.3744 \\
			20 & T\_MERG & 0.3666 \\
			21 & Quaint Critters & 0.3648 \\
			22 & TUA1 & 0.0723 \\
			23 & USTC-IAT-United & -0.4608 \\
			\hline
		\end{tabular}
	\end{table}
	
	\begin{table}[t]
		\centering
		\renewcommand\tabcolsep{6pt}
		\caption{MER-SEMI leaderboard.}
		\label{Table4}
		\begin{tabular}{c|l|c}
			\hline
			Rank & Team & Discrete $(\uparrow)$ \\
			\hline
			1 & sense-dl-lab & {0.8911} \\
			2 & SZTU-MIPS & 0.8855 \\
			3 & SUST-EiAi-Team & 0.8853 \\
			4 & Desheng & 0.8841 \\
			5 & AI4AI & 0.8811 \\
			6 & USTC-IAT-United & 0.8775 \\
			7 & SCUTer & 0.8726 \\
			8 & Voice of Soul & 0.8703 \\
			9 & Beihang University & 0.8691 \\
			10 & AIPL-BME-SEU & 0.8689 \\
			-- & \textbf{Baseline} & \textbf{0.8675} \\
			11 & ADDD & 0.8661 \\
			12 & Big Data and Intelligence Cognition & 0.8537 \\
			13 & TUA1 & 0.8507 \\
			14 & T\_MERG & 0.8486 \\
			15 & Wearing Instruments Lab & 0.0661 \\
			\hline
		\end{tabular}
	\end{table}

	\section{Participants and Outcome}
	This year's challenge attracts the registration of 76 teams from varying academic institutions. Due to the inherent class imbalance of discrete emotions (see Figure \ref{Figure2}), we choose the weighted average F-score as our evaluation metric, consistent with previous works \cite{lian2021ctnet, lian2021decn}. For dimensional emotions, we select the widely utilized mean square errors as the evaluation metric. To further evaluate comprehensive performance, we define a combined metric that incorporates both discrete and dimension predictions:
	\begin{equation}
	\mbox{metric} = \mbox{metric}_e - 0.25 * \mbox{metric}_v,
	\end{equation}
	where $\mbox{metric}_e$ and $\mbox{metric}_v$ represent the metrics for discrete emotions and valences, respectively. In MER-MULTI and MER-NOISE, participants are required to provide predictions for both discrete and dimensional emotions. Therefore, we use the combined metric for performance evaluation. In MER-SEMI, we only evaluate discrete results on the labeled subset. Therefore, we use the weighted average F-score as the evaluation metric. 
	
	For each sub-challenge, we perform an initial attempt to explore a range of multimodal features and establish a competitive baseline system\footnote{\emph{https://github.com/zeroQiaoba/MER2023-Baseline}}. To ensure reproducibility, we primarily utilize open-source pre-trained models for feature extraction and a simple yet effective multi-layer perceptron for emotion recognition. In MER-MULTI and MER-NOISE, our baseline system achieves 0.56 and 0.41 on the combined metric, respectively. In MER-SEMI, we only evaluated discrete emotions and our baseline system reaches 86.40\% on the weighted average F-score.
	
	Table \ref{Table2} $\sim$ Table \ref{Table4} show the leaderboards for the three sub-challenges. Excitingly, we witness that most teams exceed our baseline performance. The team named ``sense-dl-lab'' emerges as the winner across all three sub-challenges. Their system outperforms our baseline by 0.1405 on MER-MULTI and 0.2746 on MER-NOISE. For MER-SEMI, their system reaches 89.11\% on the evaluation metric, outperforming our baseline by 2.36\%.

	\section{Conclusions}
	This paper summarizes MER 2023, a multimodal emotion recognition challenge focused on system robustness. MER 2023 consists of three sub-challenges: (1) MER-MULTI requires participants to predict both discrete and dimensional emotions. This multi-scale labeling process can help distinguish some subtle differences in emotional states; (2) MER-NOISE simulates data corruption in real-world environments for modality robustness evaluation; (3) MER-SEMI requires participants to train more powerful classifiers using large amounts of unlabeled data. In the future, we plan to increase both labeled and unlabeled samples in our corpus. Additionally, we hope to organize a series of challenges and related workshops that bring together researchers from all over the world to discuss recent research and future directions in multimodal emotion recognition.

	\section{Acknowledgements}
	We would like to thank the members of data chairs Bin Liu (Institute of Automation, CAS), Ye Liu (Institute of Psychology, CAS) and Meng Wang (Ant Group). Meanwhile, we appreciate program committee for their valuable support: Shiguang Shan (Institute of Computing Technology, CAS), Jing Han (University of Cambridge), Liang Zhang (Institute of Psychology, CAS), Carlos Busso (University of Texas at Dallas), Rui Xia (Nanjing University of Science and Technology), Gualtiero Volpe (University of Genova), Yongwei Li (Institute of Automation, CAS), Giovanna Varni (Telecom Paris), Chi-Chun Lee (National Tsing Hua University),	Zixing Zhang (Hunan University), Xiaobai Li (University of Oulu), Heysem Kaya (Utrecht University), Jingming Zhao (Renmin University of China), Licai Sun (University of Chinese Academy of Sciences), Li Ya (Beijing University of Posts and Telecommunications), Mingyue Niu (Tianjin Normal University).
	
	This work is supported by the National Natural Science Foundation of China (NSFC) (No.61831022, No.62276259, No.62201572, No.U21B2010), Beijing Municipal Science \& Technology Commission, Administrative Commission of Zhongguancun Science Park No.Z211100004821013, Open
	Research Projects of Zhejiang Lab (No.2021KH0AB06), and CCF-Baidu Open Fund (No.OF2022025).
	\bibliographystyle{unsrt}
	\balance
	\bibliography{mybib}
\end{document}